\documentclass[runningheads]{llncs}
\usepackage[T1]{fontenc}
\usepackage{graphicx}
\usepackage{tabularx}
\usepackage{hyperref}

\begin{document}
\title{\textit{KALE-LM-Chem}: Vision and Practice Toward an AI Brain for Chemistry}
\titlerunning{KALE-LM-Chem}
%
%
\author{
Weichen Dai\inst{1,3,4}
\and
Yezeng Chen\inst{2,3}
\and
Zijie Dai\inst{1,3}
\and
Yubo Liu\inst{2,3}
\and
Zhijie Huang\inst{1,3}
\and
Yixuan Pan\inst{1,3}
\and
Baiyang Song\inst{1,3}
\and
Chengli Zhong\inst{1,3}
\and
Xinhe Li\inst{1,3}
\and
Zeyu Wang\inst{1,3}
\and
Zhuoying Feng\inst{1,3}
\and
Yi Zhou\inst{1,3}}
\authorrunning{W. Dai et al.}

\institute{
University of Science and Technology of China, Hefei, China
\and
ShanghaiTech University, Shanghai, China
\and
USTC Knowledge Computing Lab, Hefei, China
\and
State Key Laboratory of Communication Content Cognition People's Daily Online, Beijing, China}

\def\OURMODELA{{{KALE-LM-Chem}}}
\def\OURMODELB{{{KALE-LM-Chem-1.5}}}

\maketitle              
\begin{abstract}

Recent advancements in large language models (LLMs) have demonstrated strong potential for enabling domain-specific intelligence.
In this work, we present our vision for building an AI-powered chemical brain, which frames chemical intelligence around four core capabilities: information extraction, semantic parsing, knowledge-based QA, and reasoning \& planning. 
We argue that domain knowledge and logic are essential pillars for enabling such a system to assist and accelerate scientific discovery.
To initiate this effort, we introduce our first generation of large language models for chemistry: \textbf{\textit{KALE-LM-Chem}} and \textbf{\textit{KALE-LM-Chem-1.5}}, which have achieved outstanding performance in tasks related to the field of chemistry. 
We hope that our work serves as a strong starting point, helping to realize more intelligent AI and promoting the advancement of human science and technology, as well as societal development.
\footnote{Yezeng Chen and Zijie Dai are co-second authors to this paper.}
\footnote{Yi Zhou (\url{yi\_zhou@ustc.edu.cn}) is the corresponding author.}
\footnote{Models are available at \url{https://huggingface.co/USTC-KnowledgeComputingLab/Llama3-KALE-LM-Chem-8B}}
\footnote{Accepted by PRICAI 2025 as Oral.}

\end{abstract}
\section{Background}
In recent years, the rapid development of artificial intelligence (AI) technology has enabled it to achieve, and in some cases surpass, top human performance in various high-intelligence tasks. 
These include recognition in speech~\cite{chu2023qwen}, facial~\cite{alansari2023ghostfacenets}, and image~\cite{Foret2020SharpnessAwareMF}, games such as Go~\cite{Silver2017MasteringTG}, StarCraft~\cite{Arulkumaran2019AlphaStarAE}, and Dota2~\cite{Raiman2019LongTermPA}, as well as tasks related to text~\cite{Touvron2023Llama2O}, image~\cite{Kim2023ConsistencyTM}, and video generation, machine translation~\cite{Takase2021LessonsOP}, knowledge-based question answering~\cite{Yasunaga2022LinkBERTPL}, debates, and solving advanced mathematical problems~\cite{Trinh2024SolvingOG}.

Science is one of the most important fields for the application of AI. 
As the crown jewel of human civilization and the cornerstone of various industries, science is a core driver of human progress, and its development can significantly accelerate and even revolutionize many fields. 
To date, although AI has made certain progress in the scientific field, it remains far from large-scale application due to current technological limitations. 
AI primarily encompasses three stages: 
sensing/perception - cognition/thinking - decision-making/action, roughly corresponding to human subsystems such as eyes/ears/nose - brain - hands/feet.
Among these, cognition/thinking (i.e., the brain) is the core. 
Therefore, for AI in the scientific domain, constructing a scientific brain for machines is of paramount importance.

Currently, there are three main technologies for constructing scientific brains using AI, namely: specialized networks for specific problems, deep neural networks with reasoning engines, and large model based methods.

\subsubsection{Specialized Networks For Specific Problems.}
The first technology involves building specialized deep neural network models for specific problems, significantly reducing the search space. 
Google DeepMind's AlphaFold~\cite{AlphaFold2021} series is one representative work. 
This effort constructs specialized deep neural network models for protein structure prediction, greatly lowering the threshold for protein structure analysis while significantly improving its efficiency. 
Similarly, many other studies have utilized deep neural network models for scientific simulation, design, and control, vastly enhancing the efficiency of scientific research. 
For instance, DPMD~\cite{zhang2018deep}, by combining deep neural networks with high-performance computing, has dramatically expanded the capability of molecular dynamics simulations with first-principles accuracy. 
Other works have used deep learning for partial differential equation simulations~\cite{li2020fourier}, molecular property predictions~\cite{satorras2021n}, and more. 
The ABACUS-R~\cite{liu2022rotamer} adopts a data-driven strategy, paving a new path for de novo protein design. 
In the field of physics, Iten et al.~\cite{iten2020discovering} investigated how neural networks can emerge with important physical concepts, while Wu et al.~\cite{wu2019toward} constructed an AI physicist capable of abstracting theories from observational data. 
Similar research in biology includes GEARS~\cite{roohani2024predicting}, which can predict corresponding transcriptional responses to perturbations of single or multiple genes in cells.
However, these models are only applicable to certain professional fields, and each field requires custom development, leading to high development costs.

\subsubsection{Deep Neural Networks With Reasoning Engines.}
The second technology integrates deep neural networks with reasoning engines, providing new perspectives (such as auxiliary lines) for reasoning in specific domains to enhance thinking and decision-making. 
AlphaGeometry~\cite{Trinh2024SolvingOG} combines large models with symbolic engines to better solve complex problems through enhanced thinking and decision-making. 
FunSearch~\cite{RomeraParedes2023MathematicalDF} generates targeted programs to solve specific problems through the evolution of pre-trained language models and evaluators. 
Inter-GPS~\cite{Lu2021InterGPSIG} has implemented a method based on formal languages and symbolic reasoning, which shows strong interpretability in solving geometric problems. 
HAKE~\cite{Li2022HAKEAK} provides a rich space of primitives and a knowledge base, containing over 26 million primitive labels and numerous logical rules. 
FTL-LM~\cite{Lin2022FusingTC} enhances the model's application capabilities by integrating contextual information and logical rules from knowledge graphs into language models.
Similarly, these technologies also require customization and come with significant development expenses. 

\subsubsection{Large Model Based Methods.}
The third technology relies on large models for different forms of interaction. 
With the rise of ChatGPT~\cite{Achiam2023GPT4TR}, the application of large models in the scientific field has become a hot topic. 
ChemCrow~\cite{Bran2023ChemCrowAL} enhances the performance of general large models in the chemistry field through simple tool calls. 
Med-PaLM2~\cite{singhal2023towards} surpasses previous work in general medical question-answering. 
There are also studies on this technological route, such as the GeoGalactica~\cite{Lin2023GeoGalacticaAS} for earth sciences based on the general large model Galactica~\cite{taylor2022galactica}, and the ChemLLM~\cite{Zhang2024ChemLLMAC}, a scientific large model for chemistry based on InternLM~\cite{team2023internlm}. 
Thanks to the powerful generalization capabilities of LLMs, they are increasingly demonstrating their significant advantages as an AI brain.

\subsubsection{Chemistry} is a vital branch of science. 
Over decades of research and exploration, the scientific community has accumulated a vast volume of AI-ready chemical data, providing fertile ground for the development of an AI chemistry brain. 
Accordingly, in this work, we select the chemical domain as a testing ground for both theoretical exploration and practical implementation.
In the following sections, we first present our vision for building an AI-driven chemical brain, followed by a detailed description of our methodology and experimental outcomes.

\section{Vision}



As previously discussed, LLMs, empowered by pretraining on massive and diverse datasets, have demonstrated remarkable capabilities in language understanding and generalization. 
These models have been widely applied across a broad range of domains and tasks. 
Furthermore, their advanced conversational abilities make LLMs a natural foundation for constructing AI brains.
However, general-purpose LLMs alone are insufficient to meet the specialized demands of the chemistry domain. 
To develop a powerful chemistry-oriented AI brain, it is essential to further adapt these models through domain-specific training. 
This process enables the model to acquire more aligned knowledge and task-relevant capabilities for chemistry-related applications.

\subsection{Four Core Capabilities for Chemistry Tasks}

Although the field of chemistry encompasses a wide variety of tasks, we propose that these can be distilled into four fundamental capabilities: information extraction, semantic parsing, knowledge-based question answering, and reasoning \& planning. 

\textbf{Information extraction} is a crucial capability for systematically extracting structured information from raw data sources such as text, images, and other types of unstructured data ~\cite{trewartha2022quantifying,krallinger2017information,weston2019named,friedrich2020sofc,the2012integrating,liu2023generative,weston2019named,hiszpanski2020nanomaterial}. 
The goal of this process is to identify and extract key details like chemical properties, structures, reaction conditions, and experimental procedures from the data. 
This extraction forms the foundation for subsequent analysis or further computational tasks. 
Typical tasks associated with information extraction include named entity recognition, relation extraction, summarization, and image-text alignment, all of which play an essential role in transforming raw data into actionable knowledge.

\textbf{Semantic parsing} refers to the transformation of natural language descriptions into standardized, machine-readable semantic representations~\cite{Johnson_1984,Woods_1973,Zelle_Mooney_1996,Zettlemoyer_Collins_2005}. 
This process enables the system to understand and process complex chemical texts or documents in a structured manner. 
The primary objective of semantic parsing is to convert unstructured language into formats that are easily interpreted by machines for further analysis or modeling. 
Such ability can extend to generating robotic commands, potentially realizing fully automated experiments.
Typical tasks in semantic parsing include parsing and normalizing chemical reactions, rules, and synthetic pathways, which are essential for comprehending and automating chemical processes.

\textbf{Knowledge-based QA}~\cite{riloff2000rule,nassiri2023transformer} involves answering specific chemistry-related questions by utilizing embedded or external domain knowledge, such as naming conventions, properties, and reaction mechanisms. 
This capability is key for applications that require expert-level understanding and retrieval of detailed scientific information. 
Representative tasks in this area include molecular name conversion, structural descriptions, property queries, and explaining chemical mechanisms ~\cite{Kwiatkowksi_Zettlemoyer_Goldwater_Steedman_2010,Sutskever_Vinyals_Le_2014}.

\textbf{Reasoning \& planning} in the context of chemistry involves the application of domain knowledge, principles, and constraints to develop solutions to complex chemistry problems. 
Tasks in this domain include synthesis route planning, retrosynthesis and product prediction, etc., which are essential for optimizing and innovating chemical processes ~\cite{Vinyals_Kaiser_Koo_Petrov_Sutskever_Hinton_2014,Vinyals_Toshev_Bengio_Erhan_2015,Dong_Lapata_2016,Shin_Lin_Thomson_Chen_Roy_Platanios_Pauls_Klein_Eisner_Durme_2021,Scholak_Schucher_Bahdanau_2021,Roy2022BenchCLAMPAB,Jin2024AnalyzingTR}.

While conceptually distinct, they often interplay in practice. 
For instance, semantic parsing of long textual inputs may rely on information extraction to identify key elements, and complex chemistry-related questions may require reasoning over embedded or external knowledge sources before an answer can be generated.

\subsection{An Ideal AI Brain for Chemistry: Knowledge and Logic Enhanced Large Model}

Building upon the four core capabilities defined above, we envision a chemistry AI brain that can holistically assist and optimize the entire research workflow in the chemical sciences.

At the outset, leveraging its information extraction capability, the AI brain can harvest valuable data from vast volumes of literature, including the most recent publications. 
This includes theoretical insights, experimental protocols, and experiment outcomes, which are distilled into key information useful for researchers.
Next, through semantic parsing, the system converts these unstructured or semi-structured inputs into formalized semantic representations. 
These structured forms can be integrated into knowledge bases, databases, or machine-interpretable repositories, laying the groundwork for automated querying and analysis.
When presented with a novel research problem, the AI brain can retrieve relevant insights from its internal knowledge store or external sources. 
With its reasoning and planning capabilities, it incrementally constructs a solution pathway tailored to the problem.

Consider the real-world example of molecular design. 
When a chemist proposes the synthesis of a molecule with specific functionalities, the AI brain first identifies and aggregates relevant knowledge—such as functional groups or bond types—from literature or knowledge bases. 
It then associates related concepts to generate design hypotheses. 
Based on the chemical rules and prior knowledge, the model proceeds to plan feasible synthetic routes or experimental procedures. 
These procedures are then translated via semantic parsing into machine-readable instructions, which can be executed by computational simulation tools or automated laboratory robots. 
The outcomes, whether computational or experimental, are subsequently reintegrated into the system via information extraction and parsing modules, contributing to a continuously evolving body of chemical knowledge.

Throughout this closed-loop process, domain knowledge and logic (including reasoning and planning) are indispensable: the former defines the informational foundation and search space, while the latter governs the pathways of problem solving.
We therefore advocate the development of \textbf{K}nowledge \textbf{A}nd \textbf{L}ogic \textbf{E}nhanced \textbf{L}arge \textbf{M}odels (KALE-LM) as a practical and promising architecture for realizing an ideal AI brain in chemistry.
Similar to the mechanisms of human thought, large models excel in generalization, versatility, and approximate accuracy, which correspond to what is known as System 1 thinking. 
In contrast, knowledge-and-logic-based computation excels in precision, reliability, and interpretability, aligning with System 2 thinking. 
By combining these strengths, we can leverage their complementary advantages, potentially leading to the realization of strong artificial intelligence in the near future.

\section{Practice}

As previously stated, knowledge serves as the foundation of logic. 
Therefore, we first propose a training framework with a primary focus on knowledge enhancement for chemistry LLM (while we also enhance the knowledge of reasoning \& planning in this framework). 
Centered on a base model, our training paradigm targets the development of four core competencies: information extraction, semantic parsing, knowledge-based QA, and reasoning \& planning.
Our future work will further elaborate on how logic enhancement can be achieved, this constitutes the next stage of our research.

\subsection{Data Construction and Synthesis}
To comprehensively develop these four capabilities, we automatically constructed a multi-dimension training corpus from diverse public chemical data sources. 
The data sources include academic literature (e.g., ChemRxiv preprints and chemistry papers on arXiv), chemical databases (such as PubChem, USPTO, and Open Reaction Database (ORD)), and open-access chemical datasets (e.g., SMolInstruct). 

\subsubsection{Information Extraction.}
We collected millions of chemical research articles and patent documents to train the model in extracting structured chemical information from unstructured text. 
For example, the model learns to identify key entities and relations such as compound names, reaction yields, and experimental conditions from the experimental sections of scientific papers.
We began by manually annotating a small set of high-quality literature passages. 
These were then used with a teacher model via few-shot prompting to automatically annotate a large number of abstracts and experimental subsubsections, producing (text, extracted JSON) pairs. 
To ensure data quality, we applied existing chemical information extraction tools alongside pattern-based rules to verify the generated outputs and filter out clearly erroneous results.
In addition, we expanded the dataset by generating new (text, extracted JSON) pairs from structured chemical data, creating realistic and diverse examples to further enrich the training corpus. 
Summary-oriented data was also constructed by aligning paper abstracts with their corresponding full texts. 
To enhance topic diversity, we incorporated literature across various subfields such as organic chemistry and materials chemistry, ensuring the extraction task spans a broad range of domains.

\subsubsection{Semantic parsing.}
The semantic parsing data is designed to train the model to translate natural language content into structured representations. 
We constructed this type of data through the following approaches:
(1) Chemical Nomenclature Conversion: We collected aligned datasets of IUPAC names and their corresponding SMILES strings to develop the model’s bidirectional understanding of human-readable chemical names and machine-readable molecular representations.
(2) Parsing of Experimental Procedures: From textual descriptions of synthetic experiments, we extracted sequences of operations and formatted them into standardized procedural steps. 
For example, we utilized experimental records from the ORD, parsing the textual instructions into structured representations of reaction protocols.
Through these datasets, the model learns to convert complex chemical expressions into structured formats or executable commands, thereby enabling it to comprehend researchers’ intentions and support downstream automation tasks.
(3) Additional Semantic Parsing Resources: We also incorporated semantic parsing datasets from other domains, such as CONIC-10k, to further enhance the model’s ability to translate natural language into formal language.

\subsubsection{Knowledge-based QA.}
We constructed chemistry knowledge question-answer (QA) pairs to enable the model to acquire a broad understanding of chemical facts and concepts. The data sources include chemistry-related entries from Wikipedia, educational textbooks and handbooks, as well as structured content from databases such as PubChem and ChEMBL.
First, we programmatically generated fact-based QA pairs from these databases, ensuring the accuracy and authority of the answers by directly sourcing them from validated chemical repositories. 
Second, we scraped and curated questions and answers from publicly available chemistry textbooks and exam data. 
These samples span a range of question types, including fundamental concepts, experimental principles, and numerical problems.
We further employed a teacher model to generate domain-specific QA pairs automatically. 
For each subfield of chemistry, such as organic chemistry or analytical chemistry, we defined fine-grained subtopics and generated multiple questions per topic, accompanied by detailed, explanatory answers.
In addition, we incorporated existing instruction-tuning datasets such as ChemData, which include tasks like molecular property prediction, reaction prediction, and experimental analysis. 
These datasets often follow a realistic conversational format, significantly enhancing the model’s ability to perform in chemistry-focused question answering scenarios.

\subsubsection{Reasoning \& planning.}
To cultivate the model’s capabilities in reasoning and planning, we constructed a diverse set of task-specific datasets.
First, for reaction mechanism and synthesis planning, we generated tasks based on publicly available reaction databases. 
These include retrosynthesis analysis and forward synthesis prediction, for example, prompting the model to propose plausible synthetic routes for a given target molecule, or to predict the product based on specified reactants.
Second, we developed quantitative reasoning tasks, such as chemistry-related calculation problems. 
In these cases, the model is required to provide step-by-step derivations along with the final answer, thereby training its mathematical reasoning skills within a chemical context.
Third, we introduced experimental design evaluation tasks. 
We curated datasets from experimental planning questions or assessments that ask whether specific procedural steps are correct. 
For instance, given a synthetic procedure, the model may be asked to identify potentially hazardous operations or suggest improvements to enhance feasibility and safety.

\subsection{Continual Pretraining and Fine-Tuning}
To enable a smooth transition of the base model from general language proficiency to domain-specific expertise in chemistry, we designed a staged training strategy comprising two sequential phases.

\subsubsection{Phase 1: Domain-Specific Incremental Pretraining.}
Starting from a pretrained model in the general domain, we performed continual pretraining to progressively infuse domain knowledge in chemistry. 
For corpus construction, we curated a hybrid dataset combining general-domain text (also including mathematical content, code, and tool usage data) with a large volume of chemistry-related material. 
The chemical corpus covers diverse sources as described in the previous sections, including full-text journal articles, patent specifications, and database entries.

\subsubsection{Phase 2: Supervised Fine-Tuning.}
After domain-specific continual pretraining, the model acquires a strong foundation in chemical background knowledge and terminology. 
At this stage, we shift the training objective to supervised instruction tuning, leveraging our curated datasets to further optimize the model’s behavior across the four core competencies.

\section{Results}

\begin{figure*}[!ht]
    \includegraphics[width=\textwidth]{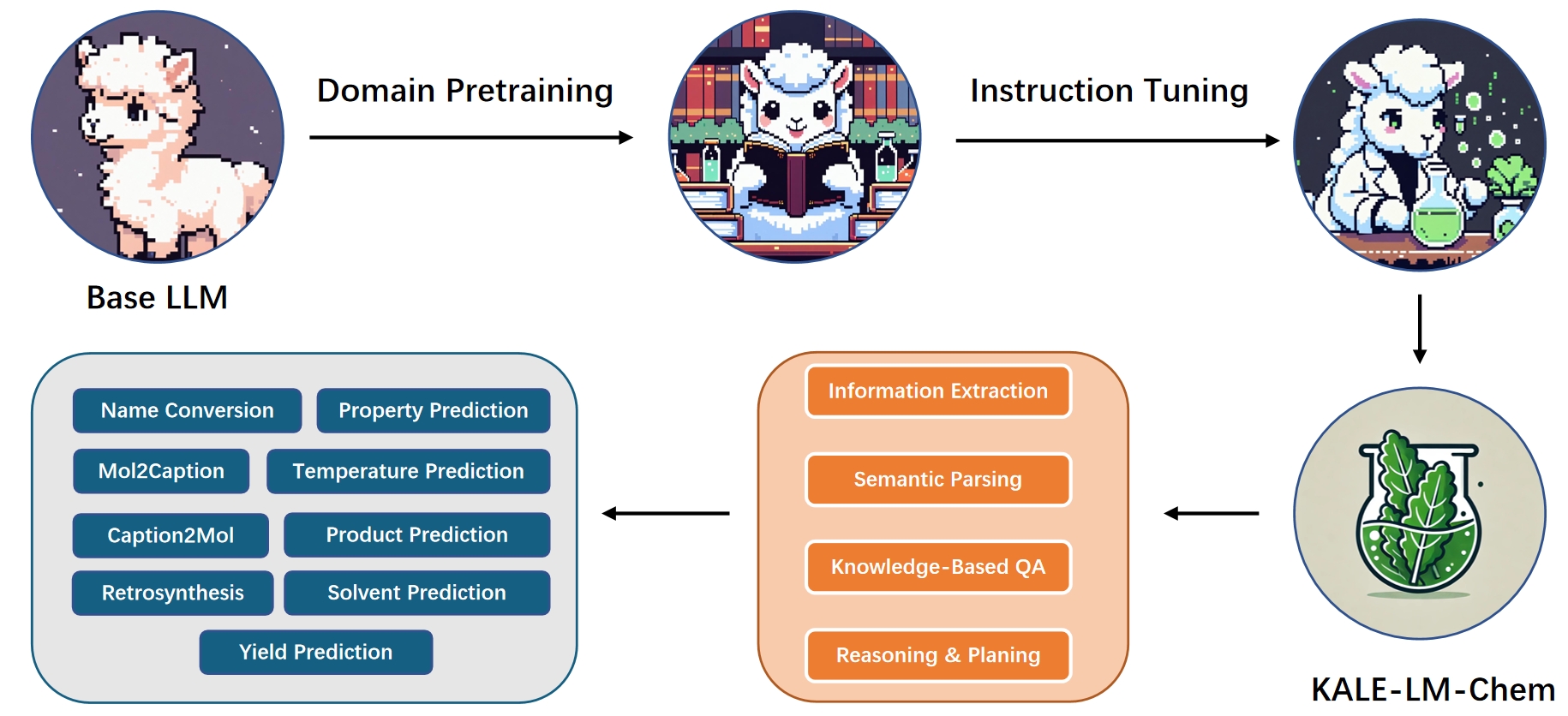}
    \caption{Training pipeline for KALE-LM-Chem.} 
\end{figure*}

\subsection{KALE-LM-Chem}


We present the first generation of our KALE-LM for chemistry: \textbf{KALE-LM-Chem} and \textbf{KALE-LM-Chem-1.5}, both of which are trained based on Llama3-8B-Instruct~\cite{grattafiori2024llama}.
The primary difference between the two lies in their parameter update strategy during the continual pretraining phase. 
KALE-LM-Chem was trained using LoRA, whereas KALE-LM-Chem-1.5 employed full-parameter activation, with all model weights updated during training. 
In the SFT stage, both models were fine-tuned in a full-parameter manner.

During continual pretraining, the maximum context length was set to 8192 tokens, while in the SFT stage, it was set to 2048 tokens. 
All training phases were conducted using the Adam optimizer and DeepSpeed ZeRO-2, distributed across multiple NVIDIA A100 80GB GPUs.

\subsection{Evaluation}

To comprehensively evaluate our models, we conducted experiments on multiple benchmark datasets and compared their performance against a range of baseline models. 
The comparison includes several powerful general-purpose language models, GPT-4o-mini (hereafter referred to as GPT-4o) and GPT-3.5-turbo (GPT-3.5), as well as leading chemistry-specific models, including LlaSMol-Mistral-7B (LlaSMol)~\cite{yu2024llasmol}, ChemDFM-13B (ChemDFM)~\cite{zhao2024chemdfm}, ChemLLM-7B-Chat (ChemLLM)~\cite{Zhang2024ChemLLMAC}, ChemLLM-7B-Chat-1.5-SFT (ChemLLM-1.5), and our base model, Llama3-8B-Instruct (Llama-3).

\begin{table}[!h]
    \centering
    \caption{Results on Chembench. \textbf{NC}: Name Conversion, \textbf{PP1}: Property Prediction, \textbf{M2C}: Molecular to Caption, \textbf{C2M}: Caption to Molecular, \textbf{PP2}: Product Prediction, \textbf{RS}: Retrosyntheis, \textbf{YP}: Yield Prediction, \textbf{TP}: Temperature Prediction, \textbf{SP}: Solvent Prediction.}
    \begin{tabular}{|c|ccccccccc|c|}
    \hline
    Models & \textbf{NC} & \textbf{PP1} & \textbf{M2C} & \textbf{C2M} & \textbf{PP2} & \textbf{RS} & \textbf{YP} & \textbf{TP} & \textbf{SP} & \textbf{Average} \\
    \hline
    GPT-3.5 & 46.93 & 56.98 & 85.28 & 38.25 & 43.67 & 42.33 & 30.33 & 42.57 & 38 & 47.15 \\
    GPT-4o & 54.82 & 65.02 & 92.64 & 52.88 & 62.67 & 52.67 & 42.33 & 24.75 & 35.67 & 53.72 \\
    \hline
    Llama-3 & 51.31 & 27.79 & 90.30 & 40.88 & 34.00 & 30.00 & 45.33 & 60.89 & 33.67 & 46.02 \\
    LlaSMol & 27.78 & 29.34 & 31.44 & 23.38 & 25.67 & 24.00 & 37.33 & 34.65 & 22.67 & 28.47 \\
    ChemDFM & 36.92 & 55.57 & 83.95 & 42.00 & 40.00 & 37.33 & 39.00 & 33.17 & 32.00 & 44.44 \\
    ChemLLM & 41.05 & 29.76 & 85.28 & 26.12 & 26.00 & 24.00 & 20.00 & 24.26 & 31.00 & 34.16 \\
    ChemLLM-1.5 & 50.06 & 49.51 & 85.28 & 38.75 & 38.00 & 26.67 & 28.33 & 31.68 & 33.67 & 42.44\\
    \hline
    \textbf{KALE} & \textbf{63.58} & \textbf{58.39} & \textbf{92.98} & \textbf{44.50} & \textbf{48.67} & \textbf{38.33} & \textbf{46.33} & \textbf{44.55} & \textbf{34.33} & \textbf{52.41} \\
    \textbf{KALE-1.5} & \textbf{61.33} & \textbf{43.44} & \textbf{90.30} & \textbf{53.62} & \textbf{72.67} & \textbf{53.67} & \textbf{46.00} & \textbf{47.03} & \textbf{45.00} & \textbf{57.01} \\
    \hline
    \end{tabular}
    \label{tab:chembench}
\end{table}

\subsubsection{ChemBench.} 
ChemBench~\cite{Zhang2024ChemLLMAC} is a comprehensive benchmark designed to evaluate the performance of AI models in chemistry-related tasks. 
It encompasses a diverse set of problems, including 
Name Conversion(NC), Property Prediction(PP1), Mol2caption(M2C), Caption2mol(C2M), Product Prediction(PP2), Retrosynthesis(RS), Yield Prediction(YP), Temperature Prediction(TP) and Solvent Prediction(SP).
This benchmark provides a rigorous assessment of model capabilities in the chemical domain, and facilitates standardized comparisons across different approaches, promoting advancements in AI-driven chemistry research.

We evaluated the performance of the LLMs on ChemBench through an LLM evaluation platform, OpenCompass~\cite{2023opencompass}, for fair comparison, and reported the results in Table ~\ref{tab:chembench}.
As shown in the table, \OURMODELA{} is significantly superior to LLM of similar scale. 
Compared to our base model Llama3-8B-Instruct, the chemical capability of \OURMODELA{} has been significantly improved.
For instance, \OURMODELA{} surpasses Llama3-8B-Instruct by a large margin in PP1 (58.39 vs. 27.79).
\OURMODELA{} also achieved higher scores in 7 out of 9 tasks compared to GPT-3.5, which is a larger model with more parameters.
Notably, \OURMODELB{} achieved the highest overall average score of 57.01, surpassing all other baseline models, including strong general-purpose models such as GPT-4o-mini (53.72).
These results highlight the effectiveness of our training framework in addressing a broad range of chemically-relevant challenges.

\begin{table*}[!h]
    \centering
    \caption{Performances on MOF information extraction. \textbf{Acc.}: Exact match accuracy, \textbf{LS}: Levenshtein distance.}
    \begin{tabular}{|c|cc|}
    \hline
    Models & \textbf{Acc.} & \textbf{LS} \\
    \hline
    GPT-3.5 & 57.75 & 73.33 \\
    GPT-4o & 62.17 & 77.92 \\
    \hline
    Llama-3 & 44.02 & 56.90 \\
    LlaSMol & 2.16 & 3.23 \\
    ChemDFM & 51.33 & 66.93 \\
    ChemLLM & 29.66 & 39.17 \\
    ChemLLM-1.5 & 14.96 & 19.61 \\
    \hline
    \textbf{KALE} & \textbf{62.89} & \textbf{76.21} \\
    \textbf{KALE-1.5} & \textbf{71.70} & \textbf{81.98} \\
    \hline
    \end{tabular}
    \label{tab:mof_ie}
\end{table*}

\subsubsection{MOF Information Extraction}

Although ChemBench is already a comprehensive benchmark for evaluating chemical language models, it does not include tasks specifically designed to assess information extraction capabilities. 
To address this gap, we conducted additional evaluations based on MOF data~\footnote{\url{https://github.com/zw-SIMM/SFTLLMs_for_ChemText_Mining}} to test the models’ performance in chemical information extraction.
We followed the method proposed in \cite{zheng2023chatgpt} to construct prompt templates and adopted two evaluation metrics: exact match accuracy and Levenshtein distance, measuring both the strict correctness and the approximate similarity between the predicted and ground-truth outputs.

As shown in Table ~\ref{tab:mof_ie}, \OURMODELB{} achieves state-of-the-art performance with an accuracy of 71.70 and an LS of 81.98, outperforming all baseline models by a significant margin. 
The previous best-performing general model, GPT-4o-mini, reaches 62.17 accuracy and 77.92 LS, while Llama3-8B-Instruct and ChemLLM-based models show considerably lower performance.
\OURMODELA{} also demonstrates strong results, achieving 62.89 accuracy and 76.21 LS, marginally outperforming GPT-4o-mini in accuracy and closely matching its LS. 
Both KALE variants exhibit stronger capability in recognizing and extracting fine-grained chemical attributes, validating their suitability for real-world information extraction tasks in chemical and materials domains.

\section{Conclusion}

In this work, we first presented our vision for an AI-powered chemical brain, which conceptualizes chemical intelligence in terms of four key capabilities. 
We also outlined how such a system could assist and accelerate scientific discovery, emphasizing that domain knowledge and logical reasoning should be regarded as its foundational pillars.
To move toward this vision, we introduced the first phase of our exploration into knowledge and logic enhanced large language models: the construction of a knowledge-enhanced chemical model. 
We detailed our training framework, including our data construction methodology and specific training strategies. 
As a result, we developed two powerful models, KALE-LM-Chem and KALE-LM-Chem-1.5.
Comprehensive evaluations across chemistry benchmarks demonstrate the effectiveness of our approach. 
Looking ahead, we plan to further investigate techniques for logic enhancement, which will complement the current knowledge-enhanced model and serve as a foundation for building a truly powerful AI-driven chemical brain.

\begin{credits}

\subsubsection{\ackname}
This work was supported by grants from the National Natural Science Foundation of China (U22B2063).
The model training was performed on the robotic AI-Scientist platform of Chinese Academy of Science.

\subsubsection{\discintname}

The authors have no competing interests to declare that are
relevant to the content of this article.
\end{credits}
%

%
%
\bibliographystyle{splncs04}
\bibliography{mybib}
%




\end{document}